\def\paperID{0000}
\def\confName{CVPR}
\def\confYear{2024}

\def\authorBlock{
    Chenyu Lin\thanks{Equal contribution}\quad
    Yusheng He\footnotemark[1]\quad
    Zhengqing Zang\footnotemark[1]\quad
    Chenwei Tang\quad
    Tao Wang\thanks{Corresponding author}\quad
    Jiancheng Lv\\
    Sichuan University\\
    {\tt\small craiglin085@gmail.com, zangzhengqing@outlook.com}
}

\newif\ifreview \newcommand{\review}{\reviewtrue}
\newif\ifarxiv 
\newif\ifcamera 
\newif\ifrebuttal 

\review 

\pdfoutput=1
\documentclass[10pt,twocolumn,letterpaper]{article}
\ifreview \usepackage[review]{cvpr} \fi
\ifarxiv \usepackage[pagenumbers]{cvpr} \fi
\ifrebuttal \usepackage[rebuttal]{cvpr} \fi
\ifcamera \usepackage{cvpr} \fi

\usepackage{graphicx}	
\usepackage{amsmath}	
\usepackage{amssymb}	
\usepackage{booktabs}
\usepackage{times}
\usepackage{microtype}
\usepackage{epsfig}
\usepackage[table,xcdraw,dvipsnames]{xcolor}
\usepackage{caption}
\usepackage{float}
\usepackage{placeins}
\usepackage{color, colortbl}
\usepackage{stfloats}
\usepackage{enumitem}
\usepackage{tabularx}
\usepackage{xstring}
\usepackage{multirow}
\usepackage{xspace}
\usepackage{url}
\usepackage{subcaption}
\usepackage{xcolor}
\usepackage[hang,flushmargin]{footmisc}

\ifcamera \usepackage[accsupp]{axessibility} \fi

% ...

\ifarxiv  \fi

\newcommand{\R}[1]{{%
    \textbf{%
        \ifstrequal{#1}{1}{\textcolor{red}{R#1}}{%
        \ifstrequal{#1}{2}{\textcolor{blue}{R#1}}{%
        \ifstrequal{#1}{3}{\textcolor{magenta}{R#1}}{%
        \ifstrequal{#1}{4}{\textcolor{teal}{R#1}}{%
                           \textcolor{cyan}{R#1}%
        }}}}%
    }%
}}

\usepackage{xr-hyper}

\makeatletter
\newcommand*{\addFileDependency}[1]{
  \typeout{(#1)}
  \@addtofilelist{#1}
  \IfFileExists{#1}{}{\typeout{No file #1.}}
}

\makeatother

\definecolor{cvprblue}{rgb}{0.21,0.49,0.74}
\usepackage[pagebackref,breaklinks,colorlinks,citecolor=cvprblue]{hyperref}
\usepackage[capitalize]{cleveref}
\crefname{section}{Sec.}{Secs.}
\crefname{table}{Table}{Tables}
\crefname{figure}{Fig.}{Figs.}

\begin{document}
\title{\textbf{VCL Challenges 2023 at ICCV 2023 Technical Report}: \\Bi-level Adaptation Method for Test-time Adaptive Object Detection}
\author{\authorBlock}
\maketitle

\begin{abstract}
This report outlines our team's participation in VCL Challenges B - Continual Test-time Adaptation, focusing on the technical details of our approach. Our primary focus is Test-time Adaptation using bi-level adaptations, encompassing image-level and detector-level adaptations. At the image level, we employ adjustable parameter-based image filters, while at the detector level, we leverage adjustable parameter-based mean teacher modules.
Ultimately, through the utilization of these bi-level adaptations, we have achieved a remarkable 38.3\% mAP on the target domain of the test set within VCL Challenges B. It is worth noting that the minimal drop in mAP, is mearly 4.2\%, and the overall performance is 32.5\% mAP.

\end{abstract}
\section{Introduction}
\label{sec:intro}
VCL Challenges are designed to encourage the development of novel methods for visual continuous learning and to provide a benchmark for evaluating the performance of different methods. 
The dataset we used in VCL Challenges B - Continual Test-time Adaptation is SHIFT~\cite{sun2022shift}, which is a multi-task driving dataset featuring the most important perception tasks under a variety of conditions and with a comprehensive sensor setup.
This report explores the technical intricacies of our approach, highlighting the strategies that have enabled us to excel in this competitive arena. Our primary focus centers on Test-time Adaptation, a critical component of computer vision systems that allows models to perform effectively across diverse domains and conditions. We have adopted a bi-level adaptation strategy, targeting image-level and detector-level adaptations. At the image level, adjustable parameter-based image filters enhance our model's ability to make accurate predictions in the face of variations in image quality, lighting, and environmental factors. Simultaneously, at the detector level, adjustable parameter-based mean teacher modules provide invaluable guidance, ensuring optimal predictions by continually updating these parameters to adapt to evolving visual data. 

Our results are impressive, with a commendable 38.3\% mAP within the target domain of the test set with just 4.2\% drop compared to 42.5\% mAP achieved within the source domain of the test set. This underscores the effectiveness of our approach. Furthermore, our overall performance shows a substantial 8.3\% mAP improvement compared to the baseline model, demonstrating the effectiveness of our bi-level adaptation strategy.


\section{Method}
\label{sec:method}

Our approach focuses on the bi-level adaptation strategy, targeting image-level and detector-level adaptations.
At the image level, image adaptive module enhances our model's ability to recognize images encountering the variations in image quality, lighting, and environmental factors. Simultaneously, at the detector level, adjustable parameter-based mean teacher module provides invaluable guidance, ensuring optimal predictions by continually updating these parameters to adapt to dynamic visual data. 

\subsection{Image Adaptive Module}
We introduce an image adaptive module with adjustable parameters to the test-time adaptation pipeline. These parameters are learned during training, allowing the filters to adapt to various testing scenarios.

\paragraph{Defogging Filter}
In order to deal with the foggy weather, we introduced the defogging filter inspired by~\cite{Liu2021ImageAdaptiveYF}.
The formulation of a hazy image is as follow~\cite{He2009SingleIH}:

\begin{equation}\label{hazy_image}
I(x)=J(x)t(x)+A(1-t(x))
\end{equation}
where $I(x)$ is the observed foggy image, $J(x)$ is the defogging image. $A$ is the global atmospheric light, $t(x)$ is the medium transmission that represents the unscattered light reaching the camera.

The simple defogging filter is controlled by two parameters $w,\alpha$. The parameter $w$ is the same as in ~\cite{Liu2021ImageAdaptiveYF}, which controls the degree of defogging:

\begin{equation}\label{defogging_w}
t(w,x)=1-w\min_C\left(\min_{\text{$y\in\Omega(x)$}}\frac{I^C(y)}{A^C}\right)
\end{equation}
while another parameter $\alpha$ is to control the global atmospheric light. To this end, we first compute the dark channel map of the haze image $I(x)$, and pick top $\alpha$ brightest pixels.Then, $A$ is estimated by averaging those $\alpha$ pixels of the corresponding position of the haze image $I(x)$.

\paragraph{Other Pixel-wise Filters}
The pixel-wise filters map the source pixel values $P_{src}=\{r_{src},g_{src},b_{src}\}$ to the target values $P_{tgt}=\{r_{tgt},g_{tgt},b_{tgt}\}$.
We implied these filters drawing on the experience of ~\cite{Liu2021ImageAdaptiveYF}. The details of these filters are showed in Table~\ref{tab_filters}.The function $En(P_{src})$ in Contrast Filter is as follows:

\begin{equation}\label{Lum}
Lum(P_{src})=0.27r_{src}+0.67g_{src}+0.06b_{src}
\end{equation}

\begin{equation}\label{EnLum}
EnLum(P_{src})=\frac{1}{2}(1-\cos{\left(\pi{\times}{Lum(P_{src}})\right)})
\end{equation}

\begin{equation}\label{En}
En(P_{src})=P_{src}{\times}\frac{EnLum(P_{src})}{Lum(P_{src})}
\end{equation}

\begin{table}[htb]\centering
\setlength{\abovecaptionskip}{5pt}
\renewcommand{\arraystretch}{1.2}
\small
\setlength{\tabcolsep}{2.0mm}{
\begin{tabular}{c c c}
\hline
Filter & Para. & Function \\
\hline
Gamma & $G$ & $P_{tgt}=P_{src}^G$ \\
Contrast & $C$ & $P_{tgt}=C{\cdot}En(P_{src}) + (1-C){\cdot}P_{src}$ \\
Exposure & E & $P_{tgt}=P_{src}{\cdot}e^{E{\cdot}\ln{2}}$ \\ 
\hline
\end{tabular}
}
\caption{The function of pixel-wise filters}
\label{tab_filters}
\end{table}

\subsection{Mix Training Strategy}
In order to simulate the harsh environment that may be encountered during testing and enable our object detector to better cope with different weather conditions, we added data augmentation for simulating foggy days and nights during training.

According to Eqs.\ref{hazy_image} and \ref{defogging_w}, the foggy image $I(x)$ for training is obtained by:

\begin{equation}\label{generate_foggy_image}
I(x)=J(x)e^{-\beta}+A(1-e^{-\beta}d(x))
\end{equation}

And $d(x)$ is defined as below:

\begin{equation}\label{generate_foggy_image_2}
d(x)=-0.04*\rho + \sqrt{max(row,col)}
\end{equation}

where $\rho$ is the Euclidean distance between the current pixel and the central pixel, row and col are the number of rows and columns of the image, respectively. Specifically, A is fixed to 0.5 during training, and $\beta=0.01*i+0.05$, where i is a random integer number range from 0 to 9 in order to control the degree of fog addition.

Specifically, A is fixed to 0.5 during training, and $\beta=0.01*i+0.05$, where i is a random integer number range from 0 to 9 in order to control the degree of fog addition.

For the simulation of night environments, we used a simple power function to increase the color depth of the RGB image. The low light image $N(x)$ is calculated by

\begin{equation}\label{generate_low_light}
N(x) = x^\eta
\end{equation}

where $\eta$ is a random float number range from 1.5 to 5.

for each image in training set, we use a 1/3 probability to load its foggy enhanced image, a 1/3 probability to load its low light enhanced image, and a 1/3 probability to load the original image under clean condition.

\subsection{Multiple Teacher Adaptation}

Under the adaptation framework of the mean teacher, the teacher model is likely to perform poorly under adverse weather conditions, resulting in poor quality pseudo labels that supervise student model learning. In this situation, the update gradient obtained by the student model backpropagation may not be beneficial to the model. 

Thus we added an additional fixed teacher model whose weights are not updated during the adaptation process. With this fixed teacher model, we can obtain more robust pseudo labels by voting on two different teacher models.


\section{Experiments}
\label{sec:experiment}

\subsection{Implementation Details}
\paragraph{Dataset}
We perform experiments on SHIFT, the largest synthetic driving video dataset~\cite{sun2022shift}, which contains 6 categories and 24 different types of discrete domain conditions which grouped into 4 main categories. According to the competetion rules, we train our model on discrete images under clean daytime condition and test it on multiple video sequences under different domain conditions. We jointly train the YOLOX backbone and proposed image filter on the training dataset.

\subsection{Quantitative Results}

The ablation study of different configurations is shown in Tab.\ref{tab:tab2}. The \textit{Detector-Adaptation} means employing multiple teacher mentioned above on the baseline mean teacher model. The \textit{Image-Adaptation} means employing the foggy and low light data augmentation during training. For the best \textit{Bi-level Adaptation} model, it uses both \textit{Detector-Adaptation} and \textit{Image-Adaptation}, and enhances the input image of the student model with foggy and low light during testing.

\begin{table*}[htb]
\centering
\setlength{\abovecaptionskip}{5pt}
\renewcommand{\arraystretch}{1.2}
\setlength{\tabcolsep}{1.0mm}{
\begin{tabular}{c|c|c|c|c|c|c}
   \hline
   \textbf{Method} & \textbf{mAP} & \textbf{mAP\_source} & \textbf{mAP\_target}
   & \textbf{mAP\_loopback} & \textbf{mAP\_drop} & \textbf{mAP\_overall}\\
   \hline  
   \rowcolor{lightgray!30} 
   \textbf{Baseline} & 38.8 & 41.0 & 33.7 & 48.7 & 7.3 & 24.2 \\
   + \textbf{\textit{Detector-Adaptation}} & 38.9 & 38.1 & 34.3 & 50.9 & 3.8 & 31.4\\
   \rowcolor{lightgray!30} 
   + \textbf{\textit{Image-Adaptation}} & 39.8 & 42.3 & 36.8 & 51.7 & 5.5 & 28.8\\
   + \textbf{\textit{Bi-level Adaptation}} & 40.9 & 42.5 & 38.3 & 50.4 & 4.2 & 32.5\\
   \hline
\end{tabular}}
\caption{Ablation of different configurations}
\label{tab:tab2}
\end{table*}


{\small
\bibliographystyle{ieeenat_fullname}
\bibliography{11_references}
}

\ifarxiv \clearpage \appendix \section{Appendix Section}
Supplementary material goes here.
 \fi

\end{document}


\title{\paperTitle}
\author{\authorBlock}
\maketitlesupplementary

\section{Appendix Section}
Supplementary material goes here.

{\small
\bibliographystyle{ieee_fullname}
\bibliography{11_references}
}